\documentclass[10pt,twocolumn,letterpaper]{article}

\usepackage[pagenumbers]{cvpr} 

\usepackage{graphicx}
\usepackage{amsmath}
\usepackage{amssymb}
\usepackage{booktabs}
\usepackage{float}

\usepackage[pagebackref,breaklinks,colorlinks]{hyperref}

\usepackage[capitalize]{cleveref}
\crefname{section}{Sec.}{Secs.}
\Crefname{section}{Section}{Sections}
\Crefname{table}{Table}{Tables}
\crefname{table}{Tab.}{Tabs.}

\newcommand\blfootnote[1]{%
  \begingroup
  \renewcommand\thefootnote{}\footnote{#1}%
  \addtocounter{footnote}{-1}%
  \endgroup
}

\usepackage{microtype}  
\usepackage{balance}  
\usepackage{overpic}
\usepackage{enumitem} 
\usepackage{overpic} 
\usepackage{color}

\definecolor{turquoise}{cmyk}{0.65,0,0.1,0.3}
\definecolor{purple}{rgb}{0.65,0,0.65}
\definecolor{dark_green}{rgb}{0, 0.5, 0}
\definecolor{orange}{rgb}{0.8, 0.6, 0.2}
\definecolor{red}{rgb}{0.8, 0.2, 0.2}
\definecolor{darkred}{rgb}{0.6, 0.1, 0.05}
\definecolor{blueish}{rgb}{0.0, 0.3, .6}
\definecolor{light_gray}{rgb}{0.7, 0.7, .7}
\definecolor{pink}{rgb}{1, 0, 1}
\definecolor{greyblue}{rgb}{0.25, 0.25, 1}

\newcommand{\Figure}[1]{Figure~\ref{fig:#1}}

\newcommand{\Table}[1]{Table~\ref{tab:#1}}

\usepackage{blindtext}

\renewcommand{\paragraph}[1]{\vspace{1em}\noindent\textbf{#1}.}
\begin{document}
\title{Pix2NeRF: Unsupervised Conditional $\pi$-GAN for Single Image to Neural Radiance Fields Translation}

\author{
    Shengqu Cai\\
    ETH Z\"urich\\
        \and
    Anton Obukhov\\
    ETH Z\"urich\\
        \and
    Dengxin Dai\\
    MPI for Informatics\\
    ETH Z\"urich\\
        \and
    Luc Van Gool \\
    ETH Z\"urich\\
    KU Leuven
}
\twocolumn[{%
  \renewcommand\twocolumn[1][]{#1}%
\maketitle
\thispagestyle{empty}
\begin{center}
  \newcommand{\teaserwidth}{0.95\textwidth}
  \vspace{-0.25in}
  \centerline{
    \includegraphics[width=\teaserwidth]{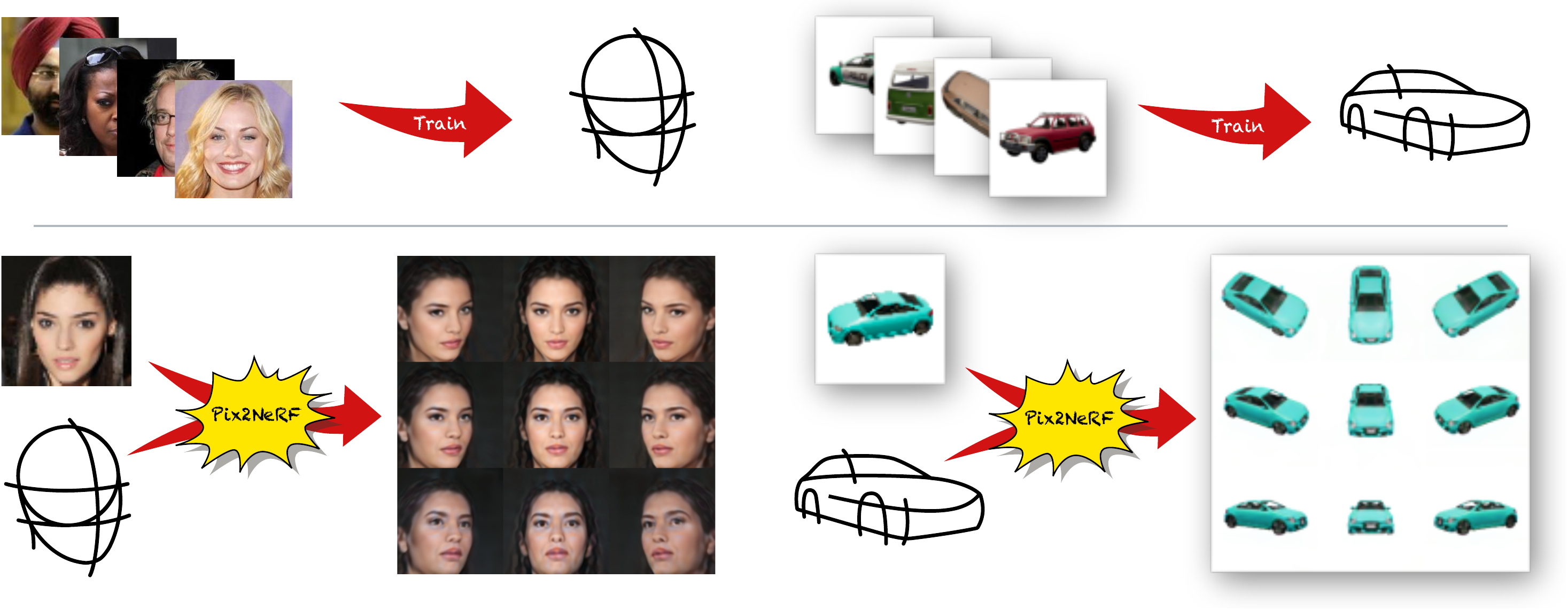}
  }
  \captionsetup{type=figure}
  \captionof{figure}{
    Overview of Pix2NeRF: We propose a method for unsupervised learning of neural representations of scenes, sharing a common pose prior. At test time, Pix2NeRF disentangles pose and content from an input image and renders novel views of the content. Top: $\pi$-GAN is trained on a dataset without pose supervision. Bottom: a trained model is conditioned on a single image to obtain pose-dependent views.
  }
  \vspace{-0.025in}
  \label{fig:teaser}
 \end{center}%
}]
\begin{abstract}
We propose a pipeline to generate Neural Radiance Fields~(NeRF) of an object or a scene of a specific class, conditioned on a single input image.
This is a challenging task, as training NeRF requires multiple views of the same scene, coupled with corresponding poses, which are hard to obtain.
Our method is based on $\pi$-GAN, a generative model for unconditional 3D-aware image synthesis, which maps random latent codes to radiance fields of a class of objects.
We jointly optimize \textbf{(1)} the $\pi$-GAN objective to utilize its high-fidelity 3D-aware generation and \textbf{(2)} a carefully designed reconstruction objective. 
The latter includes an encoder coupled with $\pi$-GAN generator to form an auto-encoder.
Unlike previous few-shot NeRF approaches, our pipeline is unsupervised, capable of being trained with independent images without 3D, multi-view, or pose supervision.
Applications of our pipeline include 3d avatar generation, object-centric novel view synthesis with a single input image, and 3d-aware super-resolution, to name a few.
\pagebreak[3]
\end{abstract}
\vspace{-1em}
\section{Introduction}
\label{sec:intro}
\blfootnote{Corresponding author: Shengqu Cai (\href{mailto://shecai@student.ethz.ch}{shecai@student.ethz.ch})}%
\blfootnote{Code~(coming soon): \href{https://github.com/HexagonPrime/Pix2NeRF}{https://github.com/HexagonPrime/Pix2NeRF}}%
Following the success of Neural Radiance Fields~(NeRF)~\cite{mildenhall2020nerf}, encoding scenes as weights of multi-layer perceptrons~(MLPs) has emerged as a promising research direction. Novel View Synthesis is an important application: given sparse sample views of a scene, the task is to synthesize novel views from unseen camera poses. NeRF addresses it by encoding color and volume density at each point of the 3D scene into a neural network and uses traditional volume rendering to compose 2D views.

While NeRF is capable of synthesizing novel views with high fidelity, it is often impractical due to being ``overfitted'' to a given scene and requiring multiple views of the scene to train.
Several follow-up works attempt to address these limitations via making NeRF generalize to new scenes. Major progress has been made in training a general NeRF capable of encoding a scene given only one or a handful of views~\cite{yu2020pixelnerf, grf2020, jang2021codenerf, wang2021ibrnet, chibane2021srf, chen2021mvsnerf}.
However, these works are designed to work well only with multi-view images during either training or both training and inference.

One reason why single-shot NeRF, or in general single-shot novel view synthesis is challenging, is the incomplete content information within a single image. For example, given a frontal image of a car, there is very little information to infer a novel view from the back directly. Bringing back the traditional inverse graphics and 3D reconstruction pipelines, \cite{Wu_2020_CVPR} addresses this issue by making an additional assumption on the symmetry of the scene to interpolate potentially missing geometry information within a single image. However, this technique is limited to scenes where symmetry can be introduced and does not tackle the general case.

Therefore, a natural follow-up question is how does a human brain address such a challenging task? One of the approaches we use unconsciously is learning a prior implicit model for object categories and mapping what we observe to the learned model. This line of thinking is already explored in prior works~\cite{yu2020pixelnerf, grf2020}. An essential part missing from these works is ensuring that novel views also meet our expectation of the object class, and due to the lack of supervision from a sole image, this is normally done via imagination.

One of the closest forms of imagination developed by the machine learning community is Generative Adversarial Networks~\cite{goodfellow2014gan}. GANs have been very successful in image synthesis and transformation. Beyond 2D, studies have shown GAN's capability of synthesizing 3D content~\cite{HoloGAN2019} from natural images. This suggests another approach to address 3D reconstruction without multi-view images via 3D GAN inversion.
Such a strategy bypasses the problem of missing information within one sole image due to GAN's adversarial training. Existing works~\cite{Pierluigi2021unsupNVS, zhang2021StyleGAN3D} utilize such a method based on HoloGAN~\cite{HoloGAN2019}, StyleGAN~\cite{zhang2021StyleGAN3D}, and others, but one of the drawbacks naturally from these 3D-aware generative models is their relatively weak 3D consistency.

With the rapid increase of NeRF~\cite{mildenhall2020nerf} popularity, corresponding generative models are also gaining attention. GRAF~\cite{Schwarz2020graf} and $\pi$-GAN~\cite{chanmonteiro2020pi-GAN} follow traditional GAN settings by mapping latent codes to category-specific radiance fields. These generative models typically have high 3D consistency due to the built-in volumetric rendering design. This observation suggests the possibility of few-shot 3D reconstruction using adversarial training and radiance fields.

In this paper, we formulate the task of translating an input image of a given category to NeRF as an end-to-end pipeline termed \textbf{Pix2NeRF} (Fig.~\ref{fig:teaser}). The method can perform novel view synthesis given a single image, without the need of pre-training, annotation, or fine-tuning. Pix2NeRF can be trained with natural images -- without explicit 3D supervision, in an end-to-end fashion. Inspired by prior works~\cite{yu2020pixelnerf, grf2020, Pierluigi2021unsupNVS}, we introduce an encoder mapping a given image to a latent space. 
We jointly optimize several objectives. First, we train $\pi$-GAN and the added encoder to map generated images back to the latent space. Second, we adapt the encoder coupled with $\pi$-GAN's generator to form a conditional GAN, trained with both adversarial and reconstruction loss. 
We show that merely doing $\pi$-GAN inversion is challenging and insufficient to complete our goal, and adaptation is important for calibrating learned representations of the encoder and generator. 
Our framework is able to instantiate NeRF in a single shot manner while naturally preserving the ability to synthesize novel views with high fidelity, comparable to state-of-the-art generative NeRF models.

\paragraph{Contributions}
\begin{itemize}
[leftmargin=*]
\setlength\itemsep{-.1em}
\item[--] We propose Pix2NeRF, the first unsupervised single-shot NeRF model, that can learn scene radiance fields from images without 3D, multi-view, or pose supervision.
\item[--] Our pipeline is the first work on conditional GAN-based NeRF, or in general, NeRF-based GAN inversion. We expect our pipeline to become a strong baseline for future works towards these research directions.
\item[--] We demonstrate the superiority of our method compared with naive GAN inversion methods and conduct an extensive ablation studies to justify our design choices.
\end{itemize}
\section{Related works}
\label{sec:related}

Our work can be classified as a category-specific 3D-aware neural novel view synthesis method, which is strongly based on NeRF~\cite{mildenhall2020nerf} and $\pi$-GAN~\cite{chanmonteiro2020pi-GAN}.

\paragraph{Neural scene representations}
The field of encoding a scene into neural networks has proven to be a promising research direction. This includes, but is not limited to: parameterizing the geometry of a scene via signed distance functions or occupancy~\cite{Park_2019_deepsdf, mescheder2019OccupancyNetworks, chen2018implicit_decoder, sitzmann2019metasdf}, encoding both geometry and appearance~\cite{sitzmann2019srns, Niemeyer2020DVR, lin2020sdfsrn, saito2019pifu}, etc. Recently, the impressive performance of Neural Radiance Fields (NeRF)~\cite{mildenhall2020nerf} has drawn extensive attention to this field. It encodes a scene as a multivariable vector-valued function $f(x,y,z,\theta,\phi)=(r,g,b,\sigma)$ approximated by MLP, where $(x,y,z)$ denotes spatial coordinates, $(\theta,\phi)$ denotes viewing direction, and $(r,g,b,\sigma)$ corresponds to color and volume density. This function is then called repeatedly by any of the volume rendering techniques to produce novel views. The outstanding performance of NeRF inspired follow-up works to extend it towards alternative settings, such as training from unconstrained images~\cite{martinbrualla2020nerfw}, training without poses~\cite{Wang21arxiv_nerfmm, meng2021gnerf}, etc.

\paragraph{NeRF-based GANs}
Following the developments of GANs and NeRFs, several works tried combining them to form generative models producing NeRFs. One of the first attempts in this direction is GRAF~\cite{Schwarz2020graf}; it performs category-specific radiance fields generation by conditioning NeRF on shape and appearance code. Following the NeRF pipeline, the generator can synthesize an image given a random code and a view direction. The generated image is passed into the discriminator together with real images, thus implementing a GAN. GRAF is an unsupervised model, since it does not require ground truth camera poses; therefore, it can be trained using "in the wild" images. 
This is done by introducing a \textit{pose prior} relative to a canonical view frame of reference, e.g., Gaussian distribution to describe head pitch and yaw relative to a front face view.
$\pi$-GAN~\cite{chanmonteiro2020pi-GAN} is similar to GRAF, but conditions on a single latent code and utilizes FiLM~\cite{ethan2017film, dumoulin2018feature-wise} SIREN~\cite{sitzmann2019siren} layers instead of simple MLPs.
More recently, several works improved synthesis quality with high resolutions~\cite{gu2021stylenerf}, better 3D shapes~\cite{xu2021gof}, and precise control~\cite{niemeyer2021GIRAFFE, zhou2021cips3d}.

\paragraph{Few-shot NeRF}
The main property of NeRFs is the ability to bake in a 3D scene into MLP weights. However, this is also a limitation since it must be retrained for each new scene, which takes a lot of time and money.
To lift this constraint, PixelNeRF~\cite{yu2020pixelnerf} and GRF~\cite{grf2020} condition MLPs on pixel-aligned features extracted by a CNN encoder. During the novel view rendering phase, 3D points along the rays are projected onto the extracted feature grid to get aligned features, then fed into an MLP with the points. More recently, CodeNeRF~\cite{jang2021codenerf} suggested training NeRF with learnable latent codes and utilizing test-time optimization to find the best latent codes~(and camera poses) given an image. However, these methods still require multi-view supervision during training, which constrains their usage in real-world settings, where multi-view datasets are challenging to collect.

Therefore, single-shot NeRF without additional supervision (e.g., 3D objects, multi-view image collections) remains an under-explored research direction. 
In this paper, we bridge this gap by incorporating an auto-encoder architecture into an existing $\pi$-GAN NeRF framework to obtain a conditional single-shot NeRF model, retaining the best properties of all components.
We note that the concurrent work~\cite{Pierluigi2021unsupNVS} shares similar ideas. The key differences are a different backbone network (HoloGAN~\cite{HoloGAN2019}) and its lack of 3D consistency, which the authors point out. 
Contrary, we utilize the newly-proposed NeRF-based GAN method called $\pi$-GAN~\cite{chanmonteiro2020pi-GAN}, which naturally provides stronger 3D consistency by design. 
We demonstrate that merely applying the approach of~\cite{Pierluigi2021unsupNVS} is insufficient to obtain an accurate mapping from image to latent space with $\pi$-GAN as a backbone.
Nevertheless, our framework can be viewed as~\cite{Pierluigi2021unsupNVS} specifically improved towards NeRF-based GAN models, or CodeNeRF~\cite{jang2021codenerf} combined with GANs.
\begin{figure*}
\begin{center}
\centering
\includegraphics[width=0.9\columnwidth, page=1, clip=true, trim = 6cm 4cm 7cm 4cm]{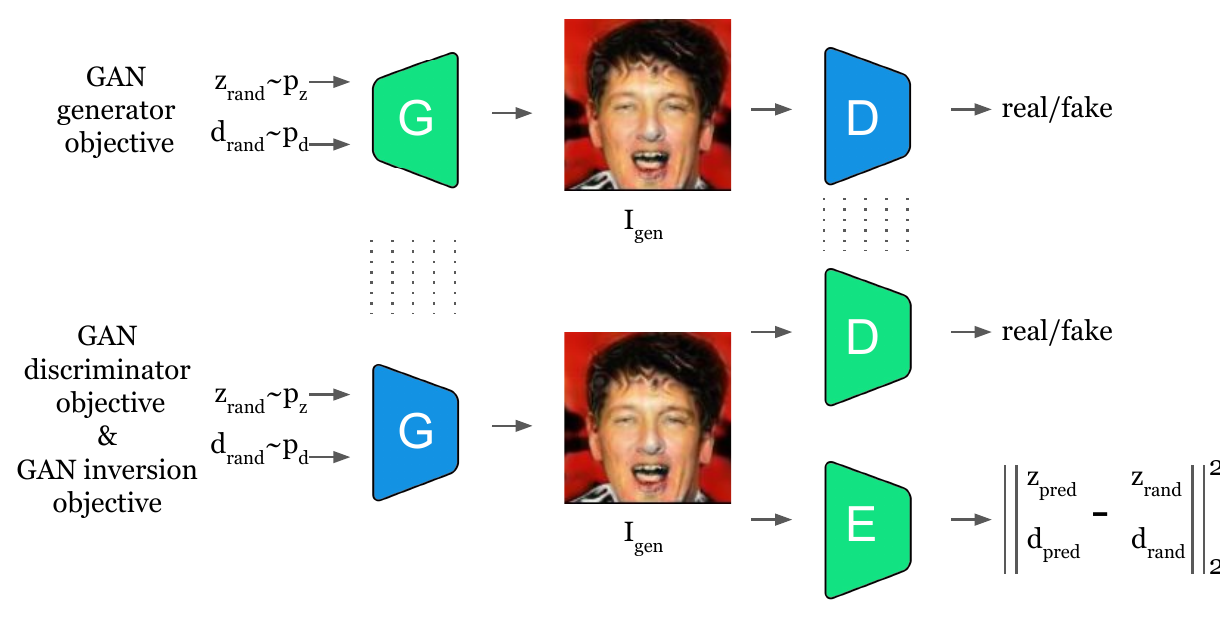}
\centering
\includegraphics[width=1.1\columnwidth, page=2, clip=true, trim = 5cm 4cm 5cm 4cm]{fig/overview_img.pdf}
\end{center}
\caption{
Overview of building blocks and objectives, used in Pix2NeRF. GAN objectives follow $\pi$-GAN~\cite{chanmonteiro2020pi-GAN} and ensure that NeRF outputs match the distribution of real images $p_\mathrm{real}$ under the latent prior $p_z$ and pose prior $p_d$. Reconstruction and GAN inversion objectives ensure calibrated latent representations, such that $E$ and $G$ can operate as an auto-encoder, similar to~\cite{Pierluigi2021unsupNVS}. The conditional adversarial objective enables learning better representations without explicit pose supervision. Legend: green - trained module, blue - frozen, gradient - warm-up.
}
\label{fig:overview}
\end{figure*}

\section{Method}
Pix2NeRF consists of three neural networks, a Generator $G$, a Discriminator $D$, together forming a Generative Adversarial Network, and an Encoder $E$ forming an auto-encoder together with $G$. 
The generator is conditioned on the output view pose $d$ and a latent code $z$, broadly describing content variations, such as color or shape. 
It employs 3D-volume rendering techniques and outputs a single parameterized scene view as RGB image $I$. 
The discriminator $D$ is a CNN, which simultaneously predicts distribution origin of the input RGB image via logit $l$ (\textit{real} -- ``in the wild'', or \textit{fake} -- generated by $G$), and the corresponding scene pose $d$.
The encoder $E$ is a CNN tasked to map an input image onto the latent manifold, learned by $G$, and at the same time predict the input's pose:
\begin{align}
\label{eq:gde}
\begin{split}
  G&: z, d \rightarrow I\\
  D&: I \rightarrow l, d\\
  E&: I \rightarrow z, d.
\end{split}
\end{align}
Functionally, Pix2NeRF extends $\pi$-GAN~\cite{chanmonteiro2020pi-GAN} with the encoder $E$ trained jointly with the GAN to allow mapping images back to the latent manifold. 
Because the encoder $E$ disentangles the content $z$ and the pose $d$ of the input $I$, content can be further used to condition the $\pi$-GAN generator $G$ and obtain novel views by varying the rendered pose $d$.

Having defined network modules, we turn to specifying the inputs and outputs of the modules. The latent code $z$ comes from a simple prior distribution $p_z$ (multivariate uniform in our case) -- it makes sampling random codes $z_\mathrm{rand}$ easy and lets us design $E$ such that it can encode any input image $I$ into some $z_\mathrm{pred}$ within the support of $p_z$. Following prior art~\cite{chanmonteiro2020pi-GAN, Schwarz2020graf}, the unsupervised setting we operate in assumes we have access to the prior distribution of poses $p_d$ of real images $I_\mathrm{real} \sim p_\mathrm{real}$ used for training. Depending on the dataset and choice of pose coordinates, it can be multivariate Gaussian with diagonal covariance (for images of faces) or uniform on a (hemi-)sphere (for images of cars). Parameters of this distribution must be known to allow easy sampling random poses $d_\mathrm{rand}$ for the generator, and that $p_d$ is representative of poses of real images $I_\mathrm{real}$.

Simply training the encoder $E$ to map an image $I$ into GAN latent space (as in Stage 1 of~\cite{Pierluigi2021unsupNVS}) simultaneously with training GAN is challenging. This is because the encoder needs to correctly map images of the same scene from different views to a single latent code. This is especially hard when these views contain variations of fine details due to occlusions. 
As seen from Eq.~\ref{eq:gde} and the design  Fig.~\ref{fig:overview}, our method disentangles latent representation of image mapped by the encoder and generator input into content $z$ and pose $d$, which undergo separate treatment. 

Given an input image, Pix2NeRF disentangles pose and content and produces a radiance field of the content, which is (1) consistent with the input under the disentangled pose and (2) consistent and realistic under different poses from $p_d$. To achieve these properties, we devise several training objectives for 
(1) generator, 
(2) discriminator, 
(3) GAN inversion, 
(4) reconstruction, and 
(5) conditional adversarial training. 

These objectives are used to compute gradients for parameters of $G$, $D$, and $E$ within a single optimization process. However, certain parts remain ``frozen'' during optimizer updates (such as $G$ during $D$ updates and vice-versa); we denote them with an asterisk in equations (e.g., $G^*$) and blue color in Fig.~\ref{fig:overview}. We empirically find that training encoder from the start has a detrimental effect on the whole pipeline and employ a warm-up strategy (denoted with green-blue transitions), explained further.

\subsection{GAN generator objective}
The generator is trained to ``fool'' the discriminator by serving it progressively realistic images. 
Pix2NeRF follows the same procedure of training the generator as $\pi$-GAN: it samples latent codes $z_\mathrm{rand} \sim p_z$ and random poses $d_\mathrm{rand} \sim p_d$ in pairs, which are then passed through the generator to obtain fake generated images:
\begin{align}
I_\mathrm{gen}=G(z_\mathrm{rand}, d_\mathrm{rand}),
\end{align}
which are further fed into the frozen discriminator:
\begin{align}
l_\mathrm{gen}, d_\mathrm{gen} &= D^*(I_\mathrm{gen}).
\end{align}
Following~\cite{chanmonteiro2020pi-GAN}, another component helpful to the stability and performance of GAN training is MSE supervision of predicted poses $d_\mathrm{gen}$ of images generated with $d_\mathrm{rand}$. It penalizes the generator if the image pose recovered by the discriminator does not correspond to the sampled pose, thus setting the goal of learning a ``canonical'' 3D space. This is especially helpful if the pose distribution of real data is noisy, such as seen in CelebA~\cite{liu2015celeba}.
\begin{align}
\begin{split}
    \mathcal{L}_\mathrm{GAN}(G)=\mathop{\mathbb{E}}_{\substack{z_\mathrm{rand} \sim p_\mathrm{z}\\ d_\mathrm{rand} \sim p_d}}\Big[&\mathrm{softplus}\left(-l_\mathrm{gen}\right) \ + \\
    &\lambda_\mathrm{pos}\left\lVert d_\mathrm{rand} -  
d_\mathrm{gen}\right\rVert_2^2\Big],
\end{split}
\end{align}
where $\lambda_\mathrm{pos}$ is a tuned weighting factor.

\subsection{GAN discriminator objective}
The discriminator is trained to distinguish between the generated fake samples and real data sampled from the dataset. Pix2NeRF follows the exact procedure of training the discriminator in $\pi$-GAN: it samples latent codes $z_\mathrm{rand} \sim p_z$ and random poses $d_\mathrm{rand} \sim p_d$ in pairs, which are then passed through the frozen generator to obtain fake generated images:
\begin{equation}
\label{eq:obj:gan:disc:Igen}
  I_\mathrm{gen}=G^*(z_\mathrm{rand}, d_\mathrm{rand}).
\end{equation}
The discriminator is then trained using these generated images $I_\mathrm{gen}$ and real images $I_\mathrm{real} \sim p_\mathrm{real}$:
\begin{align}
\begin{split}
l_\mathrm{real}, d_\mathrm{real} &= D(I_\mathrm{real}),\\
l_\mathrm{gen}, d_\mathrm{gen} &= D(I_\mathrm{gen}).
\end{split}
\end{align}

The discriminator objective modified to take into account MSE supervision over the known pose can then be formulated as follows:
\begin{align}
\begin{split}
\mathcal{L}_\mathrm{GAN}(D)=\mathop{\mathbb{E}}_{I_\mathrm{real} \sim p_\mathrm{real}} \big[&\mathrm{softplus}\left(-l_\mathrm{real}\right)\big]\ \ + \\
\mathop{\mathbb{E}}_{\substack{z_\mathrm{rand} \sim p_\mathrm{z}\\ d_\mathrm{rand} \sim p_d}} \, \Big[ &\mathrm{softplus}\left(l_\mathrm{gen}\right) \ + \\
&\lambda_\mathrm{pos} \left\lVert d_\mathrm{rand} -  
d_\mathrm{gen}\right\rVert_2^2 \Big],
\end{split}
\end{align}
where $\lambda_\mathrm{pos}$ is a tuned weighting factor.

\subsection{GAN inversion objective}
The encoder $E$ is jointly optimized with the discriminator $D$ and reuses $I_\mathrm{gen}$ computed for GAN discriminator objective Eq.~\eqref{eq:obj:gan:disc:Igen}:
\begin{align}
z_\mathrm{pred}, d_\mathrm{pred} = E(I_\mathrm{gen}).
\end{align}
This objective aims to ensure consistency between the sampled content and pose and those extracted from the generated image by the encoder. This is done using the MSE loss:
\begin{align}
\begin{split}
    \mathcal{L}_\mathrm{GAN^{-1}}(E) =
    \mathop{\mathbb{E}}_{\substack{z_\mathrm{rand} \sim p_\mathrm{z}\\ d_\mathrm{rand} \sim p_d}}\big[&
      \left\lVert z_\mathrm{pred} - z_\mathrm{rand}\right\rVert_2^2 \ \ + \\
      &\left\lVert d_\mathrm{pred} - d_\mathrm{rand}\right\rVert_2^2
    \big].
\end{split}
\end{align}
Up until now, the objectives only ensured a generative mapping from the latent space to radiance fields and some basic form of consistency to learn auto-encoder. However, our experiments show that optimizing just these three objectives does not produce a reasonable mapping. Therefore, Pix2NeRF adds two more objectives to address reconstruction quality and 3D consistency in the unsupervised setting.

\subsection{Reconstruction objective}
While the GAN inversion objective promotes consistency in latent space, nothing so far directly promotes consistency in the image space.
To this end, we condition the generator $G$ on a real image by extracting its latent code and pose prediction using the encoder, and then render its view using the predicted pose:
\begin{align}
\begin{split}
\label{eq:obj:reconst}
    z_\mathrm{pred}, d_\mathrm{pred} &= E\left(I_\mathrm{real}\right)\\
    I_\mathrm{recon} &= G\left(z_\mathrm{pred}, d_\mathrm{pred}\right).
\end{split}
\end{align}
Ideally, we expect to get back the original image. However, using MSE loss alone in the image space is known to promote structural inconsistencies and blur. 
In line with~\cite{Pierluigi2021unsupNVS}, we employ Structural Similarity Index Measure loss (SSIM~\cite{zhou2004ssim}) with weighting factor $\lambda_\mathrm{ssim}$ and a perceptual loss (VGG~\cite{Wu_2020_CVPR}) with weighting factor $\lambda_\mathrm{vgg}$. 
We can therefore aggregate the reconstruction loss as follows:
\begin{align}
\begin{split}
    \mathcal{L}_\mathrm{recon}(G, E)=
    \mathop{\mathbb{E}}_{I_\mathrm{real}\sim p_\mathrm{real}}\Big[\left\lVert I_\mathrm{recon} -  I_\mathrm{real}\right\rVert^2_2 \ + \\
    \lambda_\mathrm{ssim}\mathcal{L}_{\mathrm{ssim}}\left(I_\mathrm{recon}, I_\mathrm{real}\right) \ + \\
    \lambda_\mathrm{vgg}\mathcal{L}_{\mathrm{vgg}}\left(I_\mathrm{recon}, I_\mathrm{real}\right)\Big].
\end{split}
\end{align}

\subsection{Conditional adversarial objective}
The reconstruction objective promotes good reconstruction quality for just one view extracted by the encoder $E$. This may push the combination of networks towards either predicting trivial poses or unrealistic reconstructions for other poses from $p_d$. 
To alleviate that, we further apply an adversarial objective while conditioning the generator on an image $I_\mathrm{real}$ when it is rendered from random poses. Reusing results from Eq.~\eqref{eq:obj:reconst},
\begin{align}
\begin{split}
    l_\mathrm{cond}, d_\mathrm{cond} &= D^*(G\left(z_\mathrm{pred}, d_\mathrm{rand}\right))\\
    \mathcal{L}_\mathrm{cond}(G, E) &=
    \mathop{\mathbb{E}}_{\substack{I_\mathrm{real} \sim p_\mathrm{real}\\ d_\mathrm{rand} \sim p_d}}
    \big[\mathrm{softplus}\left(-l_\mathrm{cond}\right)\big].
\end{split}
\end{align}

\subsection{Encoder warm-up}
As pointed out in \cite{Pierluigi2021unsupNVS}, reconstruction loss may easily dominate and cause the model overfitting towards input views while losing its ability to represent 3D. We, therefore, introduce a simple ``warm-up'' strategy to counter this issue. For the first half iterations of the training protocol, we freeze the encoder while optimizing reconstruction and conditional adversarial loss and optimize only the generator for these two objectives. 
This serves as a warm-up for the generator to roughly learn the correspondence between encoder outputs and encoded images. The encoder is then unfrozen, enabling further distillation of its learned representations. 

After the warm-up stage, the encoder and generator directly form a pre-trained auto-encoder capable of producing 3D representations close to ground truth, bypassing the cumbersome early-stage reconstruction objective, which is extremely hard to balance with GAN objectives.
We show the necessity of this strategy and comparison with merely assigning a smaller weight for reconstruction loss in the ablation studies. 

\subsection{Training and Inference}
The objectives mentioned above can be trained jointly; however, we optimize them in alternative iterations due to GPU memory constraints. 
The discriminator and GAN inversion objectives are optimized upon every iteration; the GAN generator objective is optimized on even iterations; reconstruction and conditional adversarial objectives are optimized jointly during odd iterations with weighting factor $\lambda_\mathrm{recon}$:
\begin{align}
\mathcal{L}_\mathrm{odd}=\mathcal{L}_\mathrm{cond} + \lambda_\mathrm{recon}\mathcal{L}_\mathrm{recon}.
\end{align}

During the inference stage, Pix2NeRF only requires a single input image, which can be fed into the encoder $E$ and then generator $G$, coupled with arbitrarily selected poses for novel view synthesis. 
At the same time, instead of obtaining the latent code $z$ from the encoder, it is possible to sample it from the prior distribution $p_z$, to make the model  synthesize novel samples like a $\pi$-GAN.
\section{Experiments}
\label{sec:experiments}
\subsection{Evaluation}
\begin{figure*}[ht!]
\begin{center}
\includegraphics[width=0.97\textwidth, page=1, clip=true, trim = 0.75cm 0cm 7.45cm 0cm]{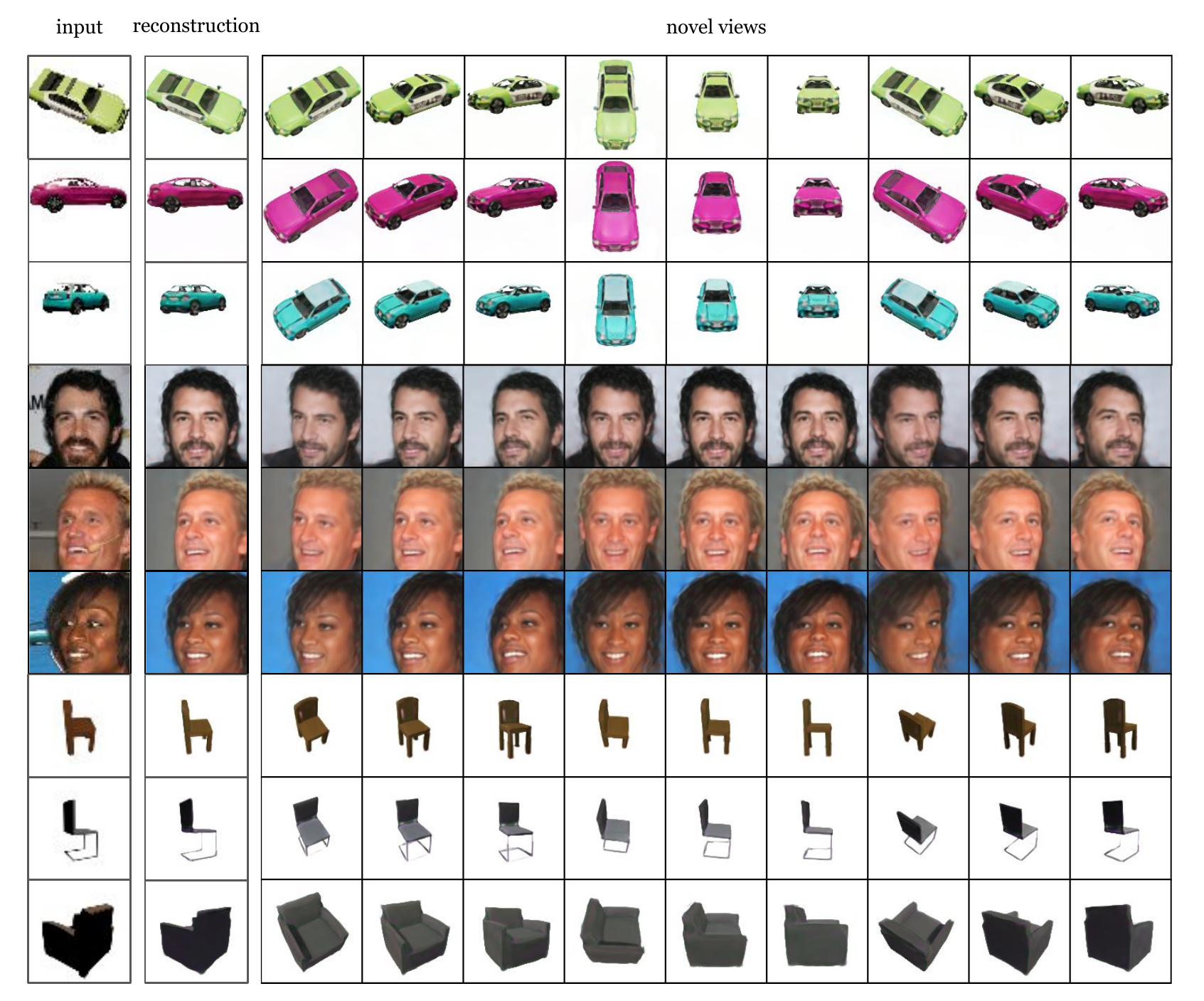}
\end{center}
\caption{
Reconstructions and novel views on CARLA~\cite{Dosovitskiy17CARLA}, CelebA~\cite{liu2015celeba}, and 
ShapeNet-SRN~\cite{angel2015shapenet, sitzmann2019srns} chairs%
. See Appendix for more results.
}
\label{fig:qualitative_results}
\end{figure*}
\paragraph{Datasets}
We train and evaluate our pipeline on several 3D datasets listed below.  
CelebA~\cite{liu2015celeba} is a dataset of over 200k images of celebrity faces. We use its ``aligned'' version and apply center cropping to keep the face area roughly. 
We hold out 8k images as the test set.
CARLA~\cite{Dosovitskiy17CARLA} contains 10k images of 16 car models rendered with Carla driving simulator with random textures.
ShapeNet-SRN is a dataset hosted by the authors of SRN~\cite{sitzmann2019srns}, from which we use the ``chairs'' split for the comparison with prior multi-view methods.
The dataset contains 50 rendered views from ShapeNet~\cite{angel2015shapenet} with Archimedean spiral camera poses for each of the 6591 instances. 
As the ShapeNet-SRN dataset does not include the lower hemisphere in its validation and test sets, we filter the training set to contain only the upper hemisphere as well.

\paragraph{Evaluation metrics}
Pix2NeRF is evaluated in two modes: unconditional, which assumes sampling directly from $p_z$ and $p_d$, and conditional, which corresponds to using $z = E(I_\mathrm{real}),\  I_\mathrm{real} \sim p_\mathrm{real}$, while still sampling from $p_d$.
For ``in the wild'' datasets, as we do not possess multi-view ground truth images, we resort to reporting generative metrics: Inception Score (IS)~\cite{salimans2016is}, Frechet Inception Distance (FID)~\cite{martin2017fid}, and Kernel Inception Distance (KID)~\cite{binkowski2021kid} with scaling factor $\times100$ following the steps of prior works~\cite{chanmonteiro2020pi-GAN, Schwarz2020graf} using the implementation~\cite{obukhov2020torchfidelity}.
To compare with multi-view-based novel view synthesis methods on Shapenet-SRN, we follow the evaluation protocols in pixelNeRF and CodeNeRF and report PSNR~(Peak Signal to Noise Ratio) and SSIM~(Structural Similarity Index Measure)~\cite{zhou2004ssim}.

\paragraph{Technical details}
We choose the latent code prior distribution $p_z$ as a multivariate uniform on $[-1, 1]$. We build our model on top of the $\pi$-GAN implementation in PyTorch~\cite{paszke2019pytorch}, re-using its released generator and discriminator architectures. We also use the discriminator architecture as the backbone of our encoder, where we add a $tanh$ at the end of the latent code head. All models are optimized with Adam~\cite{kingma2014adam} optimizer for 300k iterations, which is approximately the same computational cost to obtain a $\pi$-GAN model. CelebA~\cite{liu2015celeba} models are trained with batch size 48 on resolution 64$\times$64, where we sample 24 points per ray. We use learning rates of 2e-4, 6e-5, and 2e-4 for discriminator, generator, and encoder, respectively. 
\begin{table}[t!]
\centering
\resizebox{\linewidth}{!}{
\begin{tabular}{@{}lccc|ccc@{}}
\toprule
& \multicolumn{3}{c}{64 $\times$ 64} & \multicolumn{3}{c}{128 $\times$ 128}\\
Method & FID $\downarrow$ & KID $\downarrow$ & IS $\uparrow$ & FID $\downarrow$ & KID $\downarrow$ & IS $\uparrow$ \\
\midrule
HoloGAN~\cite{HoloGAN2019} & - & 2.87 & - & 39.7 & 2.91 & 1.89 \\
GRAF~\cite{Schwarz2020graf} & - & - & - & 41.1 & 2.29 & 2.34 \\
$\pi$-GAN~\cite{chanmonteiro2020pi-GAN} & \textbf{5.15} & \textbf{0.09} & 2.28 & \textbf{14.7} & \textbf{0.39} & \textbf{2.62} \\
\midrule
Pix2NeRF unconditional & 6.25 & 0.16 & \textbf{2.29} & 14.82 & 0.91 & 2.47 \\
Pix2NeRF conditional & 24.64 & 1.93 & 2.24 & 30.98 & 2.29 & 2.20  \\
\bottomrule
\end{tabular}
}
\caption{
Quantitative results on CelebA~\cite{liu2015celeba}.
}
\label{tab:celeba}
\end{table}%
\begin{table}[t!]
\centering
\resizebox{\linewidth}{!}{
\begin{tabular}{@{}lccc|ccc@{}}
\toprule
& \multicolumn{3}{c}{64 $\times$ 64} & \multicolumn{3}{c}{128 $\times$ 128}\\
Method & FID $\downarrow$ & KID $\downarrow$ & IS $\uparrow$ & FID $\downarrow$ & KID $\downarrow$ & IS $\uparrow$ \\
\midrule
HoloGAN~\cite{HoloGAN2019} & 134 & 9.70 & - & 67.5 & 3.95 & 3.52 \\
GRAF~\cite{Schwarz2020graf} & 30 & 0.91 & - & 41.7 & 2.43 & 3.70 \\
$\pi$-GAN~\cite{chanmonteiro2020pi-GAN} & 13.59 & \textbf{0.34} & 3.85 & 29.2 & \textbf{1.36} & 4.27 \\
\midrule
Pix2NeRF unconditional & \textbf{10.54} & 0.37 & \textbf{3.95} & \textbf{27.23} & 1.43 & \textbf{4.38} \\
Pix2NeRF conditional & 12.06 & 0.44 & 3.81 & 38.51 & 2.37 & 3.89\\
\bottomrule
\end{tabular}
}
\caption{
Quantitative results on CARLA~\cite{Dosovitskiy17CARLA}.
}
\label{tab:carla}
\end{table}%
\begin{table}[t!]
\centering
\resizebox{0.65\linewidth}{!}{
\begin{tabular}{@{}lcc@{}}
\toprule
Method & PSNR $\uparrow$ & SSIM $\uparrow$\\
\midrule
GRF*~\cite{grf2020} & 21.25 & 0.86 \\
TCO*~\cite{tatarchenko2016tco} & 21.27 & 0.88 \\
dGQN*~\cite{eslami2018dgqn} & 21.59 & 0.87 \\
ENR*~\cite{dupont2020enr} & 22.83 & - \\
SRN**~\cite{sitzmann2019srns} & 22.89 & 0.89 \\
PixelNeRF*~\cite{yu2020pixelnerf} & \textbf{23.72} & \textbf{0.91} \\
CodeNeRF**~\cite{jang2021codenerf} & 22.39 & 0.87 \\
\midrule
Pix2NeRF conditional & 18.14 & 0.84 \\
\bottomrule
\end{tabular}
}
\resizebox{0.65\linewidth}{!}{
\begin{tabular}{@{}lccc@{}}
\toprule
Method & FID $\downarrow$ & KID $\downarrow$ & IS $\uparrow$ \\
\midrule
HoloGAN~\cite{HoloGAN2019} & - & 1.54 & - \\
$\pi$-GAN~\cite{chanmonteiro2020pi-GAN} & 15.47 & 0.55 & \textbf{4.62} \\
\midrule
Pix2NeRF unconditional & \textbf{14.31} & \textbf{0.51} & \textbf{4.62} \\
Pix2NeRF conditional & 17.55 & 0.59 & 4.36 \\
\bottomrule
\end{tabular}
}
\caption{
Quantitative results on ShapeNet-SRN~\cite{angel2015shapenet, sitzmann2019srns} chairs. Top: reconstruction metrics ($128\times128$). Bottom: generative metrics ($64\times64$). Legend: * -- requires multi-view training data; ** -- requires multi-view training data and test time optimization.
} 
\label{tab:srn_chairs}
\end{table}%
For all other models, we utilized $\pi$-GAN~\cite{chanmonteiro2020pi-GAN}'s progressive training strategy, starting with training on resolution 32$\times$32 with learning rates 4e-5, 4e-4, and 4e-4 for generator, discriminator, and encoder, respectively, with 96 sampled points per ray. We increase to resolution 64$\times$64 with learning rates 2e-5, 2e-4, and 2e-4 for generator, discriminator, and encoder, respectively, and sample 72 points per ray after 50k iterations. We empirically set $\lambda_\mathrm{recon}=5$,  $\lambda_\mathrm{ssim}=1$ and $\lambda_\mathrm{vgg}=1$ for all datasets. For CelebA~\cite{liu2015celeba}, we follow~\cite{chanmonteiro2020pi-GAN} and set $\lambda_\mathrm{pos}=15$. For CARLA~\cite{Dosovitskiy17CARLA} and ShapeNet-SRN~\cite{angel2015shapenet, sitzmann2019srns}, we set $\lambda_\mathrm{pos}=0$ as we do not observe significant difference. We use $|z|=512$ for CelebA~\cite{liu2015celeba} and $|z|=256$ for CARLA~\cite{Dosovitskiy17CARLA} and Shapenet-SRN~\cite{angel2015shapenet, sitzmann2019srns}.

\paragraph{Quantitative results}
We show the evaluation on CelebA~\cite{liu2015celeba} and CARLA~\cite{Dosovitskiy17CARLA} in Tables~\ref{tab:celeba} and \ref{tab:carla} respectively. We also show evaluation with the same generative metrics on ShapeNet-SRN in \Table{srn_chairs}~(bottom).
We observe that even though our model's conditional synthesis is not as good as our backbone $\pi$-GAN (especially on CelebA), it is on par with other prior 3D view generation methods~\cite{Schwarz2020graf, HoloGAN2019}. 

Since we do not explicitly enforce prior distribution $p_z$ on the encoded samples $E(I_\mathrm{real})$ from $p_\mathrm{real}$, the image of $p_\mathrm{real}$ resulting from the encoder mapping may occupy a small portion in $p_z$.
Thus, conditioning on $p_\mathrm{real}$ naturally leads to a smaller variation in samples from $p_z$, and hence, smaller diversity of NeRF outputs. 
For this reason, directly sampling randomly from $p_z$ (unconditionally) achieves better performance as measured by the generative metrics. 
Additionally, our generator outperforms $\pi$-GAN on most metrics on CARLA~\cite{Dosovitskiy17CARLA} and ShapeNet-SRN~\cite{angel2015shapenet, sitzmann2019srns}.
Results on CelebA~\cite{liu2015celeba} are less consistent due to dataset noise (background, geometry, pose noise, artifacts, \etc), encouraging GANs to converge towards the mean as a trade-off to variations.
These observations can be related to manifold learning~\cite{du2021manifoldlearning}, where we enforce the existence of a latent code for each real image in the train set.

We compare our method with other single-image 3D inference methods in \Table{srn_chairs} on ShapeNet-SRN~\cite{angel2015shapenet, sitzmann2019srns} in 128 $\times$ 128 resolution. Since our model assumes a strictly-spherical camera parameterization model, which does not correspond well to the ground truth poses of ShapeNet-SRN~\cite{angel2015shapenet, sitzmann2019srns}, we use our encoder to extract poses from the images. 

Despite being generative, unsupervised, and not requiring test time optimization in contrast to all other methods, our model's performance does not drop much below the competition. Considering that other models were trained on 128, while our models were trained on 64 $\times$ 64 but rendered at 128 $\times$ 128 resolution, we observe a super-resolution effect.

\paragraph{Qualitative results}
We show some qualitative results of our model's performance on CARLA~\cite{Dosovitskiy17CARLA} and CelebA~\cite{liu2015celeba} in Fig.~\ref{fig:qualitative_results}. We can see that our model can synthesize novel views with good quality while existing few-shot NeRF methods~\cite{yu2020pixelnerf, grf2020, jang2021codenerf} are not able to train on these ``in the wild'' datasets due to the lack of multi-view supervision. Our model can also produce decent 3D representations even under extreme poses and artifacts~(see row 5).

\subsection{Ablation studies}
\begin{figure}[t]
\begin{center}
\includegraphics[width=\columnwidth, page=1, clip=true, trim = 1cm 5.3cm 11.5cm 0cm]{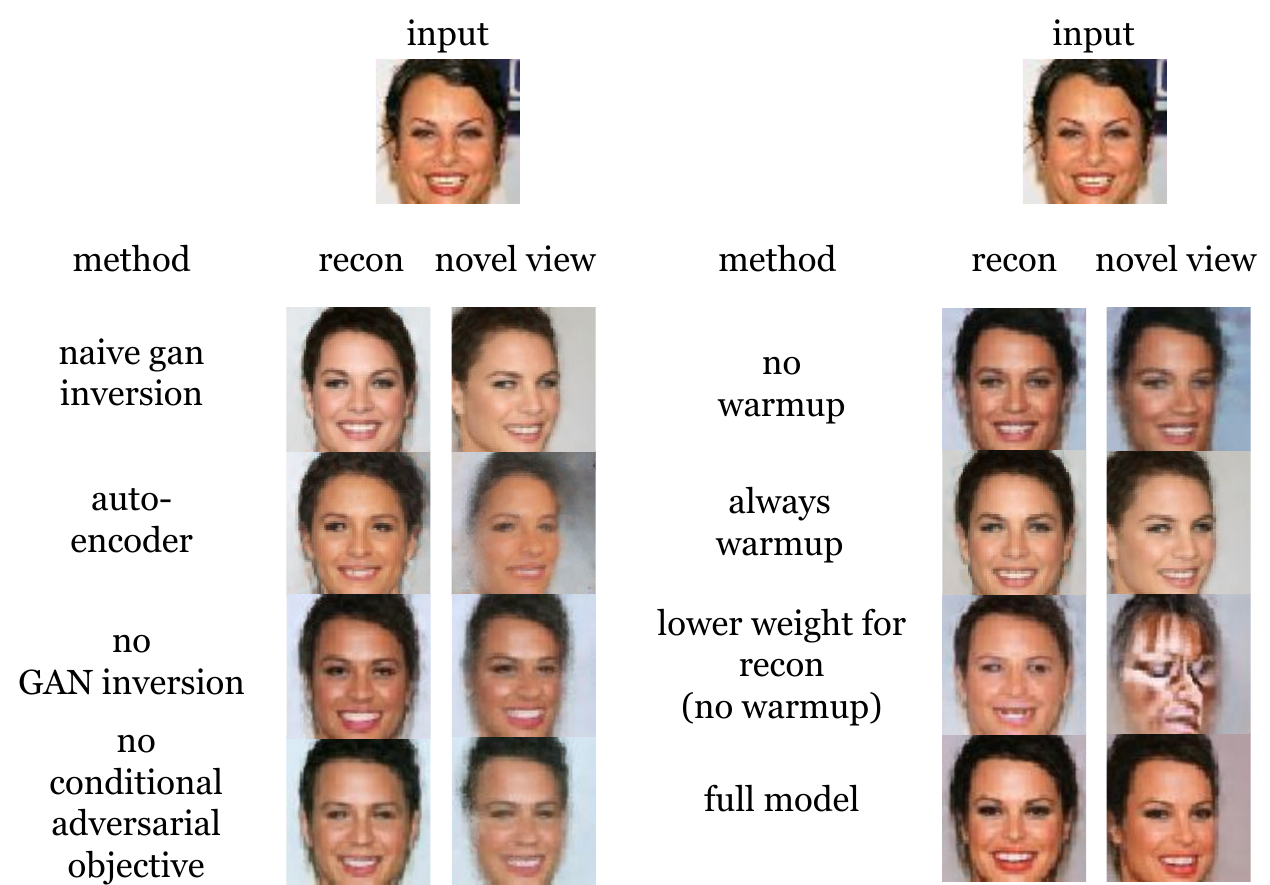}
\end{center}
\caption{
Qualitative results of ablation studies, obtained with an image from the test split of CelebA~\cite{liu2015celeba}. $\lambda_\mathrm{recon}$ is set to 1 for lower reconstruction weights instead of the warm-up ablation. See Appendix for results obtained by using other $\lambda_\mathrm{recon}$ values.}
\label{fig:ablations}
\end{figure}

We perform a thorough ablation study to verify our design choices by removing the key components one by one and training models under identical settings as the full model. 
Qualitative results for the following ablations are in Fig.~\ref{fig:ablations}; refer to Appendix for the corresponding quantitative results.

\paragraph{Naive GAN inversion} 
We compare Pix2NeRF with naive GAN inversion: having a pre-trained GAN, we freeze its weights and train an encoder to map images to their corresponding latent codes. 
The results show that the encoder can learn an approximate mapping from images to latent code. However, due to the lack of joint distillation, the reconstruction is off from the input image.

\paragraph{Auto-encoder} 
Another potential approach is to utilize $\pi$-GAN's architecture as an auto-encoder, in which the latent space is dropped from the pipeline and training the reconstruction and conditional adversarial objectives only.
Under this setting, while the reconstruction achieves decent quality, we can observe visible 3D inconsistency,
suggesting difficulty of optimization with the remaining objectives.

\paragraph{No GAN inversion}
We proceed with ablations by removing the GAN inversion step from the pipeline. The visual results turn out to be blurry and uncanny compared with full settings. One possible explanation is that this step is a connection between $\pi$-GAN training and reconstruction, which significantly affects the overall performance.

\paragraph{No conditional adversarial objective}
We further deactivate the conditional adversarial loss and retrain the model. As a result, the renderings become incomplete and have clear visual artifacts. In addition, 3D consistency degrades significantly, which justifies this objective in the given setting.

\paragraph{Warm-up}
To verify the effect of the warm-up strategy, we train three separate models and compare their performances: without warm-up, without unfreezing encoder~(always warm-up), and assigning a lower weight for reconstruction instead of the warm-up.
Without the warm-up strategy, the model tends to overfit the input view and cannot produce meaningful content from novel poses.
If we only use the warm-up strategy and never unfreeze the encoder, the distillation is relatively weak, which results in few fine details.
With lower reconstruction weight instead of the warm-up, the balance between reconstruction and adversarial objective is missing, resulting in mode collapse for novel view synthesis.
\section{Conclusions}

In this paper, we introduced Pix2NeRF, a novel unsupervised single-shot framework capable of translating an input image of a scene into a neural radiance field (NeRF), thereby performing single-shot novel view synthesis. The key idea of Pix2NeRF is to utilize generative NeRF models to interpolate missing geometry information. This is accomplished by jointly training an encoder that maps images to a latent space, which disentangles content and pose, and the generative NeRF model while keeping these two parts dependent on each other. 
Pix2NeRF can go beyond the auto-encoder setting and perform novel scene generation by sampling random content and pose and passing through the generator. 
Our framework demonstrates high reconstruction quality and 3D consistency, on par and better than previous works.

\paragraph{Limitations and future work}
The current setting in consideration is limited to one category per dataset and cannot directly generalize beyond the chosen category. 
Alternative research directions include local conditional fields similar to PixelNeRF~\cite{yu2020pixelnerf} and GRF~\cite{grf2020}, which can generalize to unseen categories, multi-instance, and even real-world scenes.
Being a general framework, Pix2NeRF is not limited to using $\pi$-GAN as its backbone. Newer generative NeRF models, e.g. EG3D~\cite{Chan2021EG3D} could potentially achieve better visual quality.
Additionally, architecture search, especially with respect to the encoder remains a challenging problem. Utilizing more mature encoder architectures from 2D GAN feed-forward inversion literature, e.g. pixel2style2pixel~\cite{richardson2021pixel2style2pixel}, could potentially improve the performance of Pix2NeRF significantly.

\paragraph{Ethical consideration}
As with most modern conditional generative models, Pix2NeRF can be misused by generating content to spread misinformation or perform targeted attacks. The growing popularity of deepfake celebrity accounts in social media suggests that new use cases, markets, and novel ways of monetizing this kind of data will follow.

\pagebreak

{
    \balance
    \small
    \bibliographystyle{ieee_fullname}
    \bibliography{macros,main}
}

\appendix


\twocolumn[
\centering
\Large
\textbf{Pix2NeRF: Unsupervised Conditional $\pi$-GAN for Single Image to Neural Radiance Fields Translation} \\
\vspace{0.5em}Supplementary Material \\
\vspace{1.0em}
] 
\appendix
\setcounter{page}{1}

\section{Additional qualitative results}
\begin{figure*}[ht!]
\begin{center}
\includegraphics[width=0.99\textwidth, page=1, clip=true, trim = 0cm 5cm 0cm 0cm]{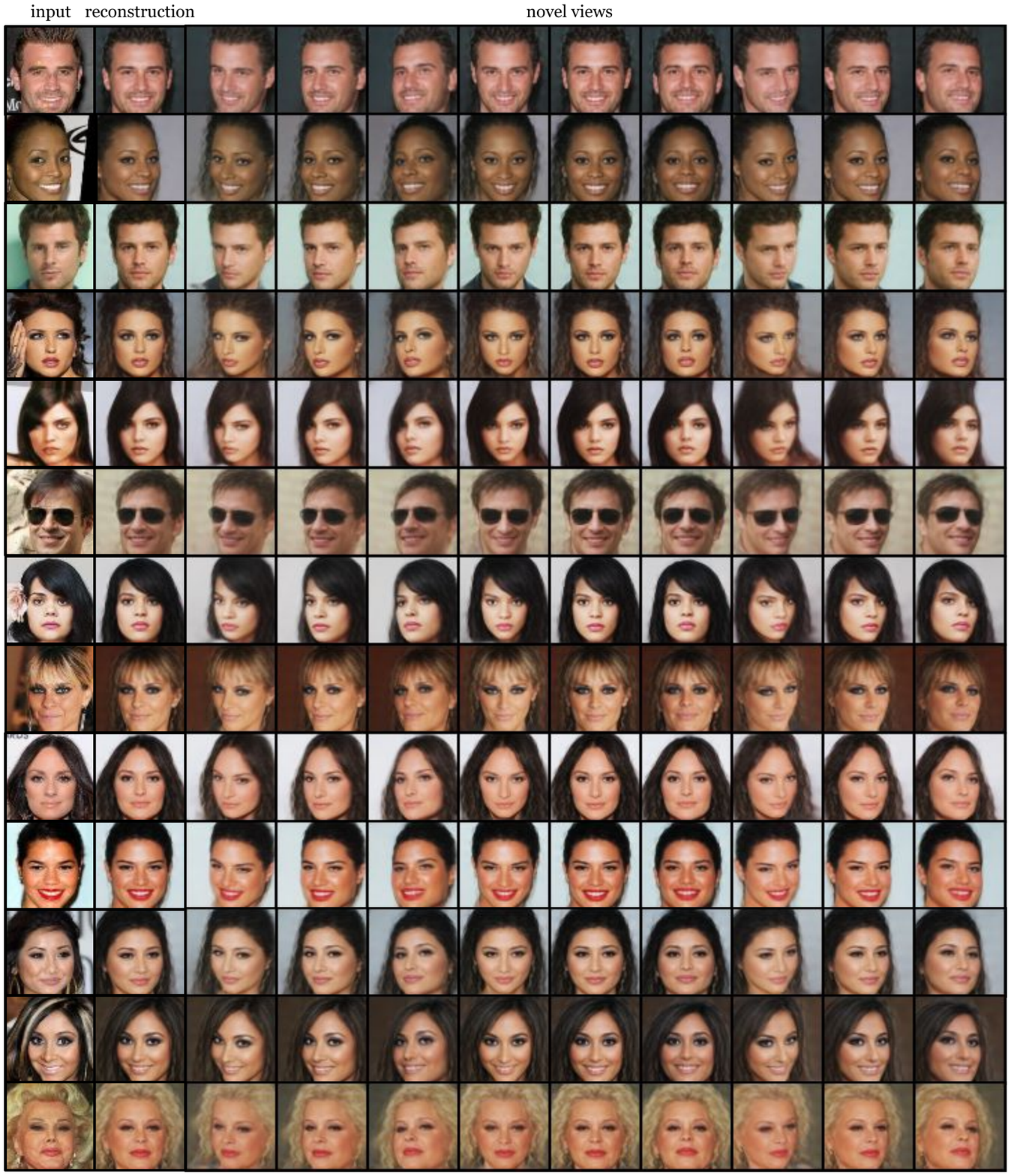}
\end{center}
\caption{
Further reconstructions and novel views on CelebA~\cite{liu2015celeba}.
}
\label{fig:celeba_qualitative_results}
\end{figure*}

\begin{figure*}[ht!]
\begin{center}
\includegraphics[width=0.95\textwidth, page=1, clip=true, trim = 0cm 4cm 0cm 0cm]{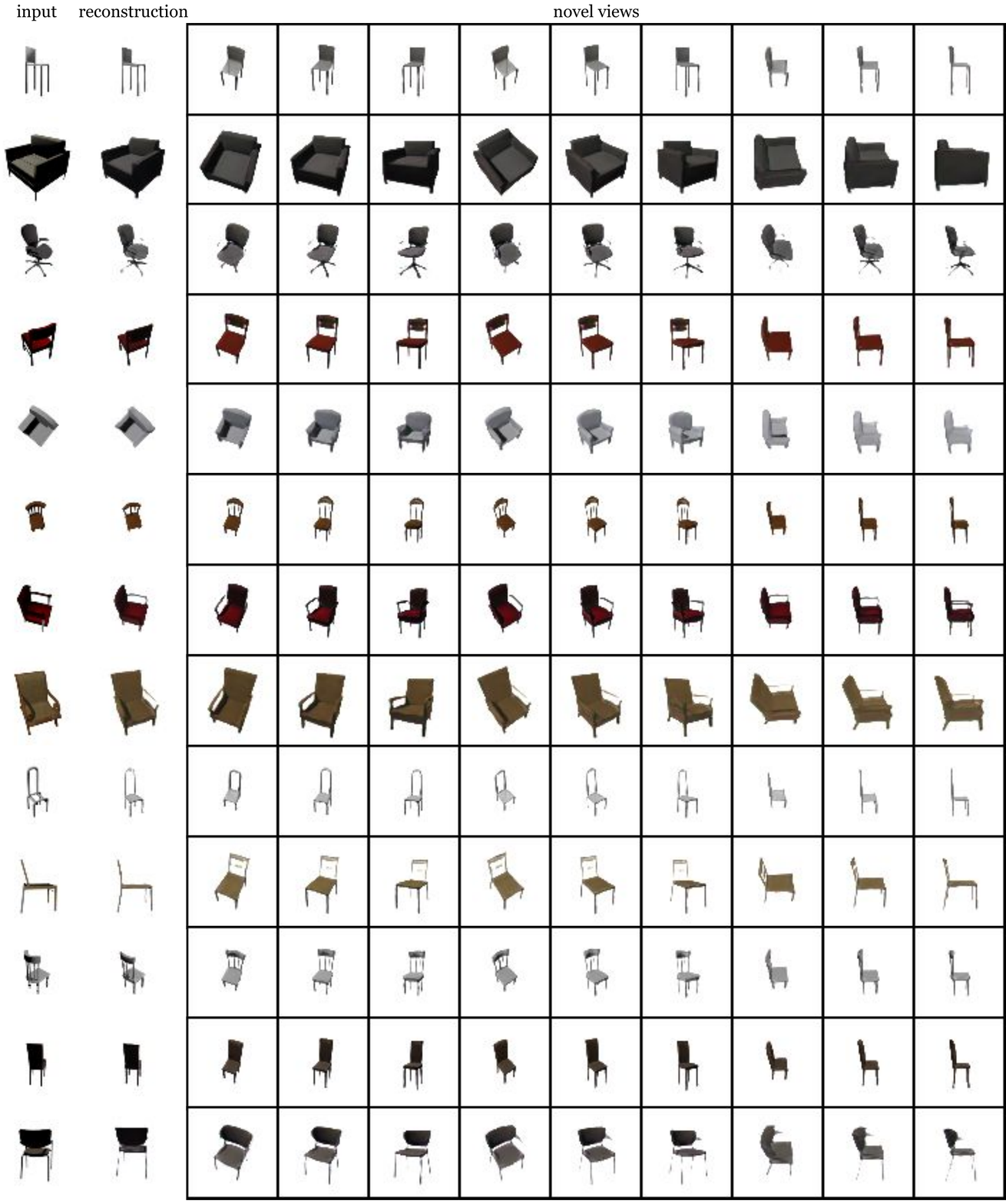}
\end{center}
\caption{
Further reconstructions and novel views on ShapeNet-SRN~\cite{angel2015shapenet, sitzmann2019srns}.
}
\label{fig:srnchairs_qualitative_results}
\end{figure*}

\begin{figure*}[ht!]
\begin{center}
\includegraphics[width=0.99\textwidth, page=1, clip=true, trim = 0cm 4.5cm 0cm 0cm]{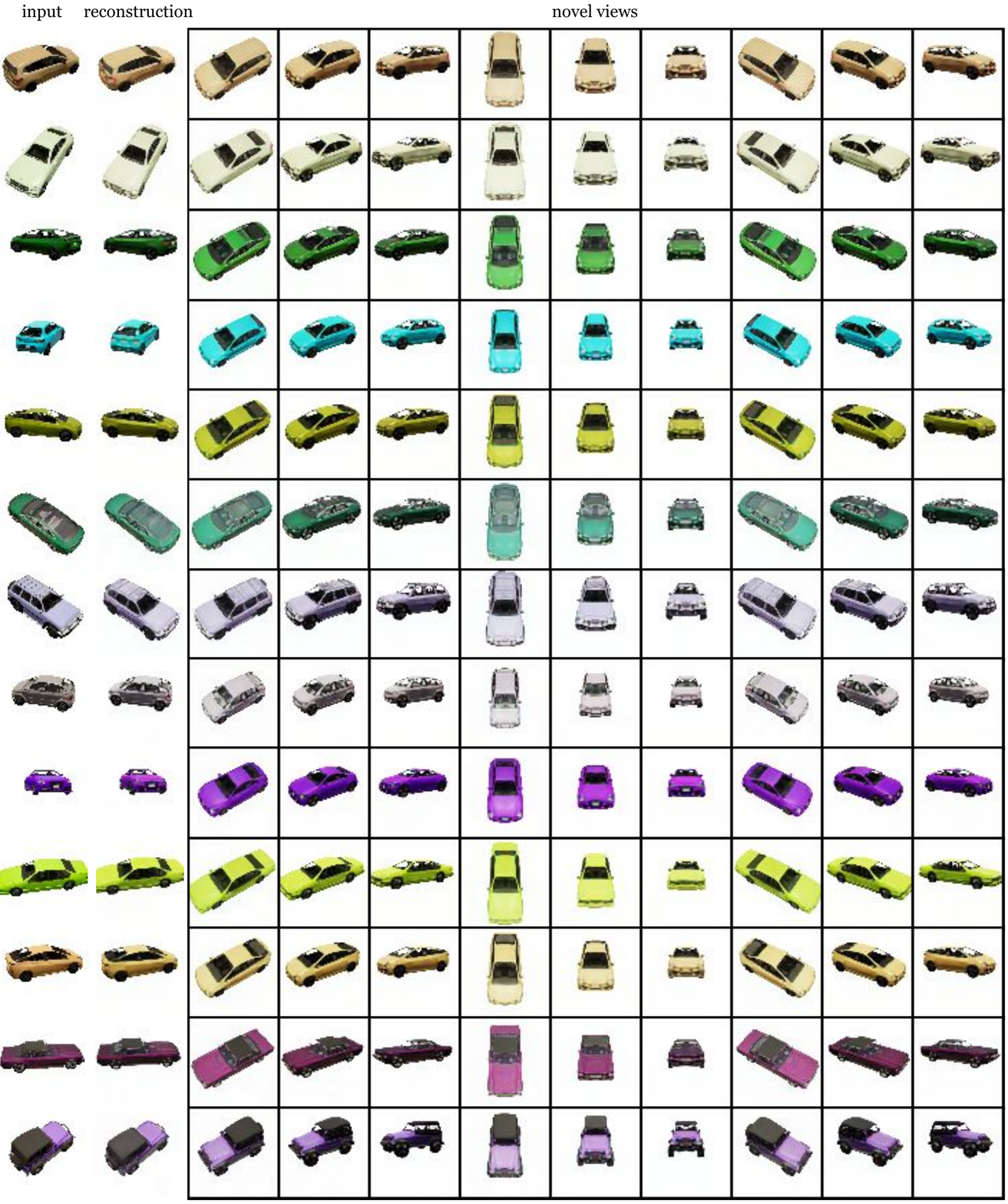}
\end{center}
\caption{
Further reconstructions and novel views on CARLA~\cite{Dosovitskiy17CARLA}.
}
\label{fig:carla_qualitative_results}
\end{figure*}

We demonstrate additional qualitative results achieved by Pix2NeRF on three datasets: CelebA~\cite{liu2015celeba}, Shapenet-SRN chairs~\cite{angel2015shapenet, sitzmann2019srns}, and CARLA~\cite{Dosovitskiy17CARLA} in Figures~\ref{fig:celeba_qualitative_results}, \ref{fig:srnchairs_qualitative_results}, and \ref{fig:carla_qualitative_results} respectively.

\section{Additional quantitative results}
\Table{srnchairs_additional_quantitative} provides additional quantitative results on ShapeNet-SRN~\cite{angel2015shapenet, sitzmann2019srns} with generative metrics computed on 128 $\times$ 128 resolution, and reconstruction metrics computed on 64 $\times$ 64 resolution. We do not report PSNR and SSIM for CelebA~\cite{liu2015celeba} as there is no ground truth novel views.

\section{Additional ablation study}
We provide quantitative results of each ablation study on CelebA~\cite{liu2015celeba} and Shapenet-SRN~\cite{angel2015shapenet, sitzmann2019srns} to further verify our design choices. As in the ablation study in our main paper, we report FID~\cite{martin2017fid}, KID~\cite{binkowski2021kid} and IS~\cite{salimans2016is} for CelebA~\cite{liu2015celeba}, and additionally report PSNR and SSIM~\cite{zhou2004ssim} on Shapenet-SRN~\cite{angel2015shapenet, lin2020sdfsrn}. We measure results after inference on resolution 64 $\times$ 64.
We show quantitative ablation results in \Table{quantitative_ablations}.
Legend: \textbf{A} -- naive GAN inversion; \textbf{B} -- auto-encoder; \textbf{C} -- no GAN inversion; \textbf{D} -- no conditional adversarial objective; \textbf{E} -- no warm-up; \textbf{F} -- always warm-up; \textbf{G}, \textbf{H}, \textbf{I}, \textbf{J} -- lower weights for reconstruction instead of warm-up, with $\lambda_\mathrm{recon} = 1, 0.1, 0.01, 0.001$ respectively. Note that since the encoder output is not enforced to strictly follow $p_z$, naive GAN inversion (stage 1 in ~\cite{Pierluigi2021unsupNVS}) failed completely due to bad initialization. We therefore use a ``warmed-up'' version of the generator trained for 300k iterations.

\section{Input reconstruction and hybrid optimization}
\begin{figure}[t!]%
\setlength{\belowcaptionskip}{-1pt}%
\centering%
\includegraphics[width=0.99\linewidth, trim={2.8cm 10cm 7.75cm 0cm},clip]{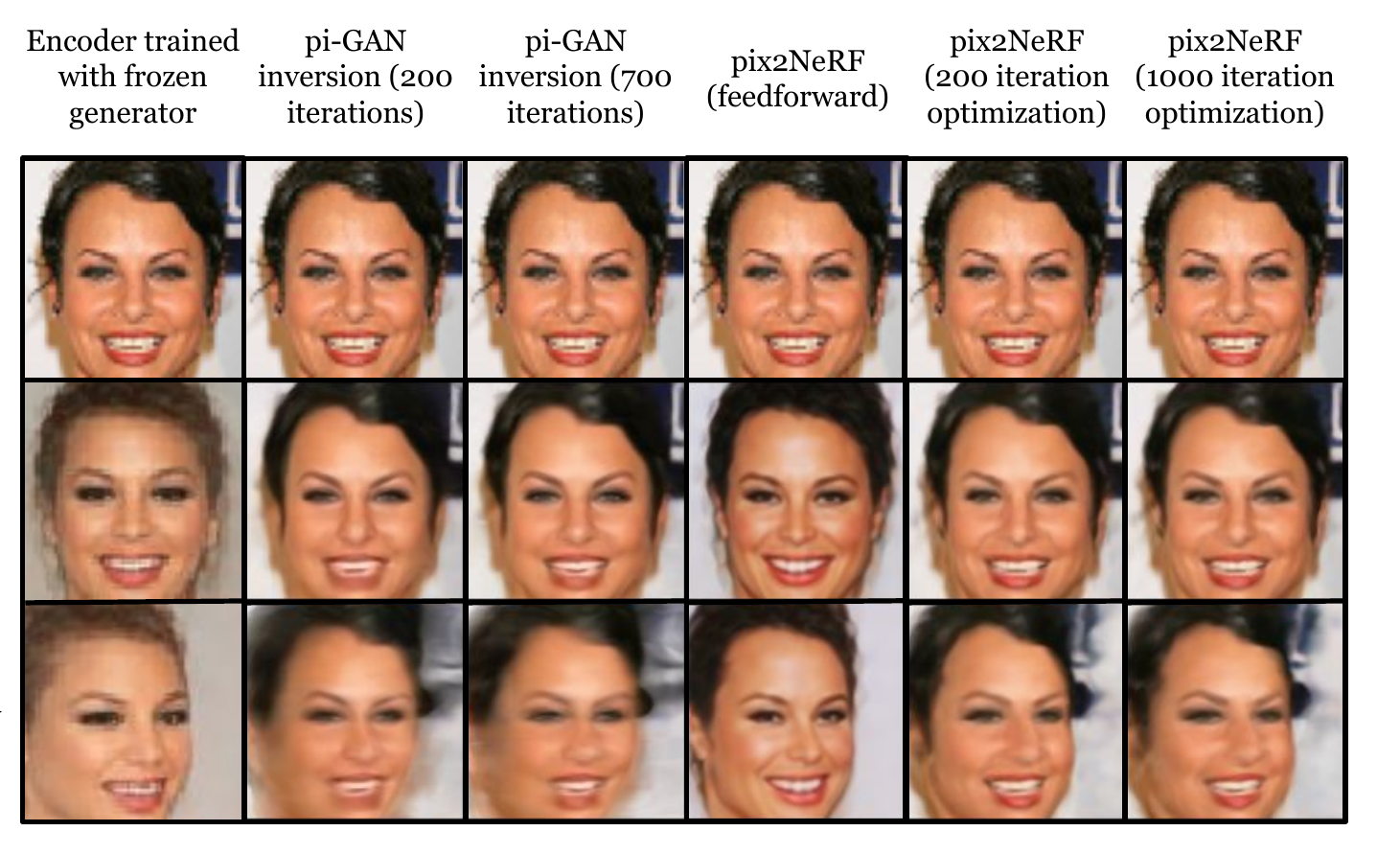}%
\caption[]{
Qualitative comparison on CelebA. 
Top -- input, middle -- reconstruction, bottom -- novel view synthesis.
}
\label{fig:comparison_pigan_inversion}
\end{figure}

We ran extra ablations and summarized our model performance by providing both input reconstruction~(cols 2,3) and novel view synthesis~(cols 4,5,6) results in Tab.~\ref{table:pigan_nvs}~(row 4).
We show $\pi$-GAN latent optimization on an input image for 700 iterations, as recommended by its authors in row 3. 
Note that it requires time-consuming per-instance optimization due to the NeRF's rendering mechanism. 
Additionally, we use the Pix2NeRF encoder's output as a starting point and perform latent optimization with a frozen Pix2NeRF generator for only 200 iterations, shown in row 5. 
A qualitative comparison is shown in Fig.~\ref{fig:comparison_pigan_inversion}. 
Note that our model does not overfit the input view even with 1000 iterations of input view optimization~(row 6), while $\pi$-GAN shows strong artifacts and requires a search for the optimal number of iterations.

\section{Necessity of generator distilling}
We trained the encoder with a pretrained frozen $\pi$-GAN generator using all the losses. As can be seen from the results in Tab.~\ref{table:pigan_nvs} Row 1, the model struggles to capture details accurately without fine-tuning the generator jointly.

\section{Linear interpolation}
\begin{figure*}[ht!]
\begin{center}
\includegraphics[width=0.99\textwidth, page=1, clip=true, trim = 0cm 13cm 0cm 0cm]{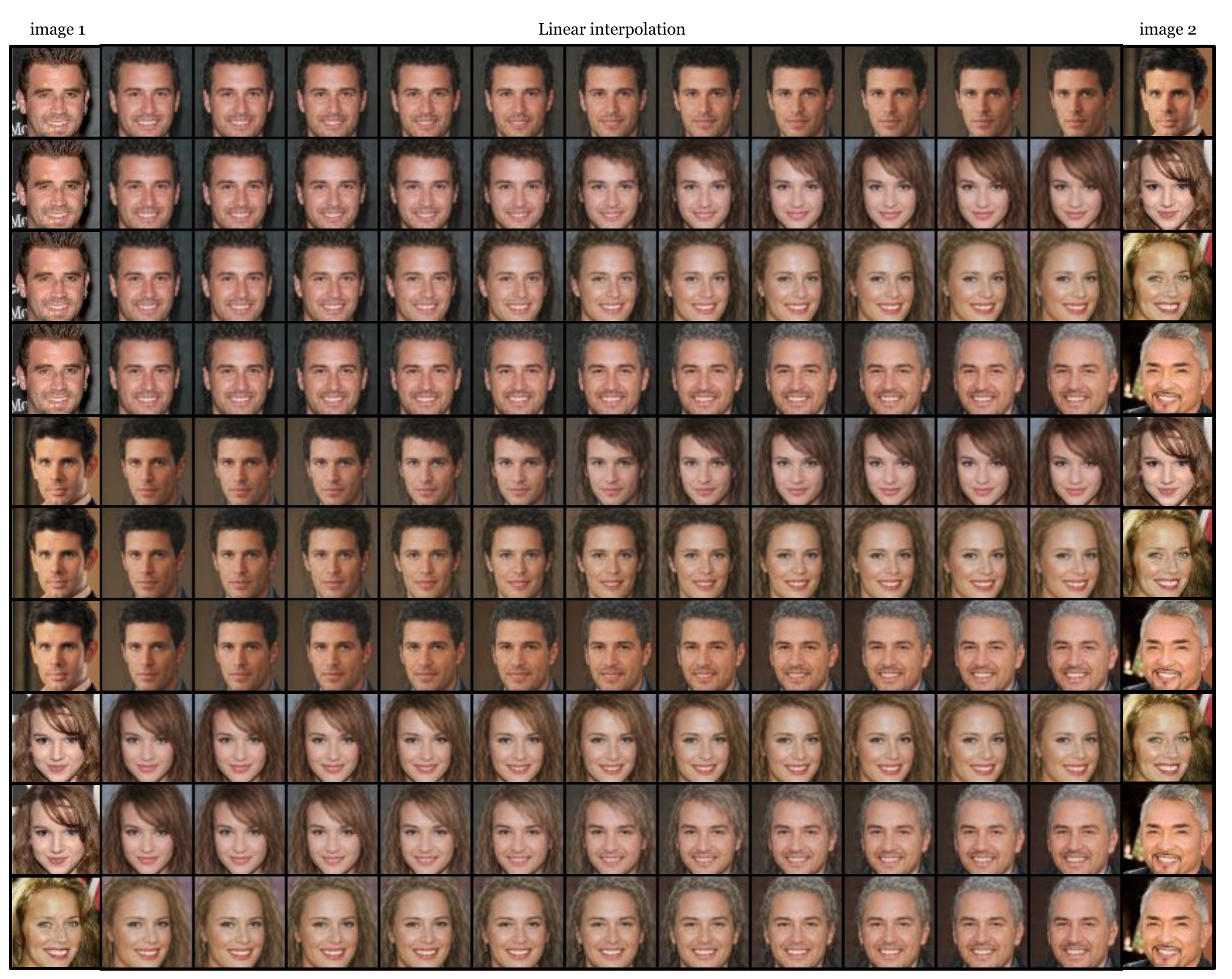}
\end{center}
\caption{
Linear interpolation on CelebA~\cite{liu2015celeba}.
}
\label{fig:linear_interpolation}
\end{figure*}

We interpolate novel views between two different input images by predicting their corresponding latent codes and poses, then applying linear interpolation to get the intermediate codes and poses. We show the results interpolating five images in \Figure{linear_interpolation}.

\begin{table}[t!]
\centering
\resizebox{0.99\linewidth}{!}{
\begin{tabular}{@{}lccc|cc@{}}
\toprule
& \multicolumn{3}{c}{128 $\times$ 128} & \multicolumn{2}{c}{64 $\times$ 64}\\
Method & FID $\downarrow$ & KID $\downarrow$ & IS $\uparrow$ & PSNR $\uparrow$ & SSIM $\uparrow$ \\
\midrule
Pix2NeRF unconditional & 26.45 & 1.18 & 4.39  & - & - \\
Pix2NeRF conditional & 26.81 & 1.23 & 4.27 & 18.75 & 0.82 \\
\bottomrule
\end{tabular}
}
\caption{
Additional quantitative results on ShapeNet-SRN~\cite{angel2015shapenet, sitzmann2019srns}.
}
\label{tab:srnchairs_additional_quantitative}
\end{table}%
\begin{table}[t!]
\centering
\resizebox{\linewidth}{!}{
\begin{tabular}{@{}lccc|ccccc@{}}
\toprule
& \multicolumn{3}{c}{CelebA 64 $\times$ 64} & \multicolumn{5}{c}{ShapeNet-SRN 64 $\times$ 64}\\
Method & FID $\downarrow$ & KID $\downarrow$ & IS $\uparrow$ & FID $\downarrow$ & KID $\downarrow$ & IS $\uparrow$ & PSNR $\uparrow$ & SSIM $\uparrow$ \\
\midrule
\textbf{A} & 28.90 & 2.99 & 1.62 & 34.01 & 1.73 & 3.65 & 15.91 & 0.71\\
\textbf{B} & 43.19 & 2.84 & 1.33 & 43.06 & 2.49 & 2.92 & 16.27 & 0.71\\
\textbf{C} & 39.42 & 3.07 & 1.65 & 41.47 & 2.80 & 2.96 & 15.14 & 0.68\\
\textbf{D} & 33.92 & 2.84 & 1.87 & 35.72 & 1.74 & 3.75 & 16.81 & 0.77\\
\textbf{E} & 31.31 & 2.75 & 1.95 & 21.67 & 0.89 & 4.35 & 18.03 & 0.79\\
\textbf{F} & 39.86 & 3.18 & 1.73 & 27.70 & 1.22 & 4.09 & 16.98 & 0.77\\
\textbf{G} & 73.52 & 7.47 & 1.91 & 27.10 & 1.31 & 4.26 & 17.77 & 0.79\\
\textbf{H} & 73.03 & 7.08 & 1.97 & 41.11 & 2.27 & 3.34 & 14.98 & 0.74\\
\textbf{I} & 140.25 & 16.33 & 1.79 & 184.10 & 17.19 & 2.55 & 10.95 & 0.59\\
\textbf{J} & 168.59 & 18.89 & 1.50 & 266.64 & 30.29 & 1.98 & 10.28 & 0.47\\
\midrule
Full & \textbf{24.64} & \textbf{1.93} & \textbf{2.24} & \textbf{17.55} & \textbf{0.59} & \textbf{4.36}  & \textbf{18.75} & \textbf{0.82}\\
\bottomrule
\end{tabular}
}
\caption{
Quantitative results of ablation study on CelebA~\cite{liu2015celeba} and ShapeNet-SRN~\cite{angel2015shapenet, sitzmann2019srns}. ``Full'' denotes Pix2NeRF conditional setup.
}
\label{tab:quantitative_ablations}
\end{table}%
\begin{table}[t!]
\caption{
Input view reconstruction~(PSNR, SSIM) on a test set, and novel view synthesis~(FID, KID$\times$100, IS).
}
\label{table:pigan_nvs}
\normalsize
\centering
\resizebox{1.0\linewidth}{!}{
\begin{tabular}{l c c c c c}
\toprule
Method & PSNR$\uparrow$ & SSIM$\uparrow$ & FID$\downarrow$ & KID$\downarrow$ & IS$\uparrow$ \\
\midrule
Pix2NeRF $E$ + frozen $\pi$-GAN $G$ & 13.04 & 0.46 & 28.25 & 2.97 & 1.52\\
$\pi$-GAN optimization~(200 iterations)  & 23.42 & 0.80 & 16.09 & 0.83 & 2.10 \\
$\pi$-GAN optimization~(700 iterations)  & 24.21 & 0.82 & 17.14 & 0.72 & 2.14 \\
Pix2NeRF~(feed-forward)  & 17.95 & 0.67 & 24.82 & 1.93 & 2.21 \\
Pix2NeRF~(200 iterations)  & 27.12 & 0.89 & 12.86 & 0.64 & 2.27\\
Pix2NeRF~(1000 iterations)  & \textbf{27.73} & \textbf{0.90} & \textbf{12.01} & \textbf{0.62} & \textbf{2.30}\\
\bottomrule
\end{tabular}
}
\end{table}

\section{Limitations and failure cases}
\begin{figure*}[ht!]
\begin{center}
\includegraphics[width=0.25\textwidth, page=1, clip=true, trim = 0cm 0cm 0cm 0cm]{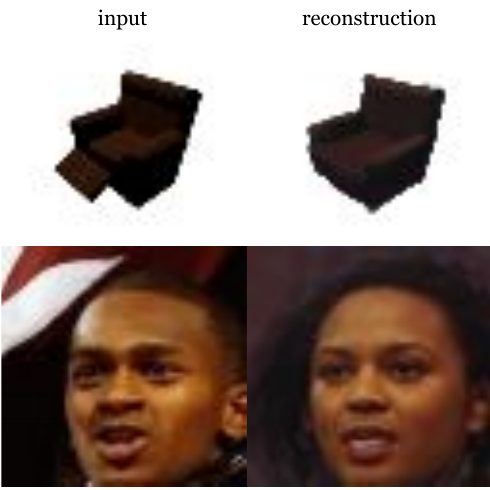}
\end{center}
\caption{
Failure cases on CelebA~\cite{liu2015celeba} and ShapeNet-SRN~\cite{angel2015shapenet, sitzmann2019srns}.}
\label{fig:failure_cases}
\end{figure*}

Despite training on images without pose or 3D supervision, Pix2NeRF can reconstruct objects from a single image and achieve decent quality. However, the methodology of using an encoder to encode an entire image into a single latent code is quite challenging, especially when the dataset is noisy, such as CelebA~\cite{liu2015celeba}. Pix2NeRF cannot always capture fine details accurately. We observe failure cases when the input is out-of-distribution relative to that of the training set $p_\mathrm{real}$, as shown in \Figure{failure_cases}. It might be possible to improve these hard cases by introducing pixel-wise features instead of (or, in addition to) the global latent code, as done in PixelNeRF~\cite{yu2020pixelnerf} and GRF~\cite{grf2020}. 


\end{document}